\documentclass[runningheads]{llncs}
\usepackage{graphicx}
\usepackage{comment}
\usepackage{amsmath,amssymb} 
\usepackage{color}
\usepackage{multirow}
\usepackage{booktabs}
\usepackage[pagebackref=true,breaklinks=true,letterpaper=true,colorlinks,bookmarks=false]{hyperref}

\usepackage{xspace}
\makeatletter
\DeclareRobustCommand\onedot{\futurelet\@let@token\@onedot}
\def\@onedot{\ifx\@let@token.\else.\null\fi\xspace}

\def\eg{\emph{e.g}\onedot} 
\def\ie{\emph{i.e}\onedot}

\def\etal{\emph{et al}\onedot}
\makeatother

\begin{document}
\pagestyle{headings}
\mainmatter
\def\ECCVSubNumber{121}  

\title{Learning Object Relation Graph and\\ Tentative Policy for Visual Navigation} 

%
\author{Heming Du\inst{1} \and
Xin Yu\inst{1,2} \and
Liang Zheng\inst{1}}
\authorrunning{Du et al.}
%
\institute{Australian National University \\
\email{\{heming.du,liang.zheng\}@anu.edu.au} \and
University of Technology Sydney \\
\email{xin.yu@uts.edu.au}}
\maketitle

\begin{abstract}
Target-driven visual navigation aims at navigating an agent towards a given target based on the observation of the agent. In this task, it is critical to learn informative visual representation and robust navigation policy.
Aiming to improve these two components, this paper proposes three complementary techniques, object relation graph (ORG), trial-driven imitation learning (IL), and a memory-augmented tentative policy network (TPN). ORG improves visual representation learning by integrating object relationships, including category closeness and spatial correlations, \emph{e.g.,} a TV usually co-occurs with a remote spatially. 
Both Trial-driven IL and TPN underlie robust navigation policy, instructing the agent to escape from deadlock states, such as looping or being stuck. Specifically, trial-driven IL is a type of supervision used in policy network training, while TPN, mimicking the IL supervision in unseen environment, is applied in testing. 
Experiment in the artificial environment AI2-Thor validates that each of the techniques is effective. When combined, the techniques bring significantly improvement over baseline methods in navigation effectiveness and efficiency in unseen environments.
We report 22.8\% and 23.5\% increase in success rate and Success weighted by Path Length (SPL), respectively.
The code is available at \url{https://github.com/xiaobaishu0097/ECCV-VN.git}.
\keywords{Graph, imitation learning, tentative policy learning, visual navigation}
\end{abstract}

\section{Introduction}
\begin{figure}[t]
    \centering
    \includegraphics[width=\textwidth]{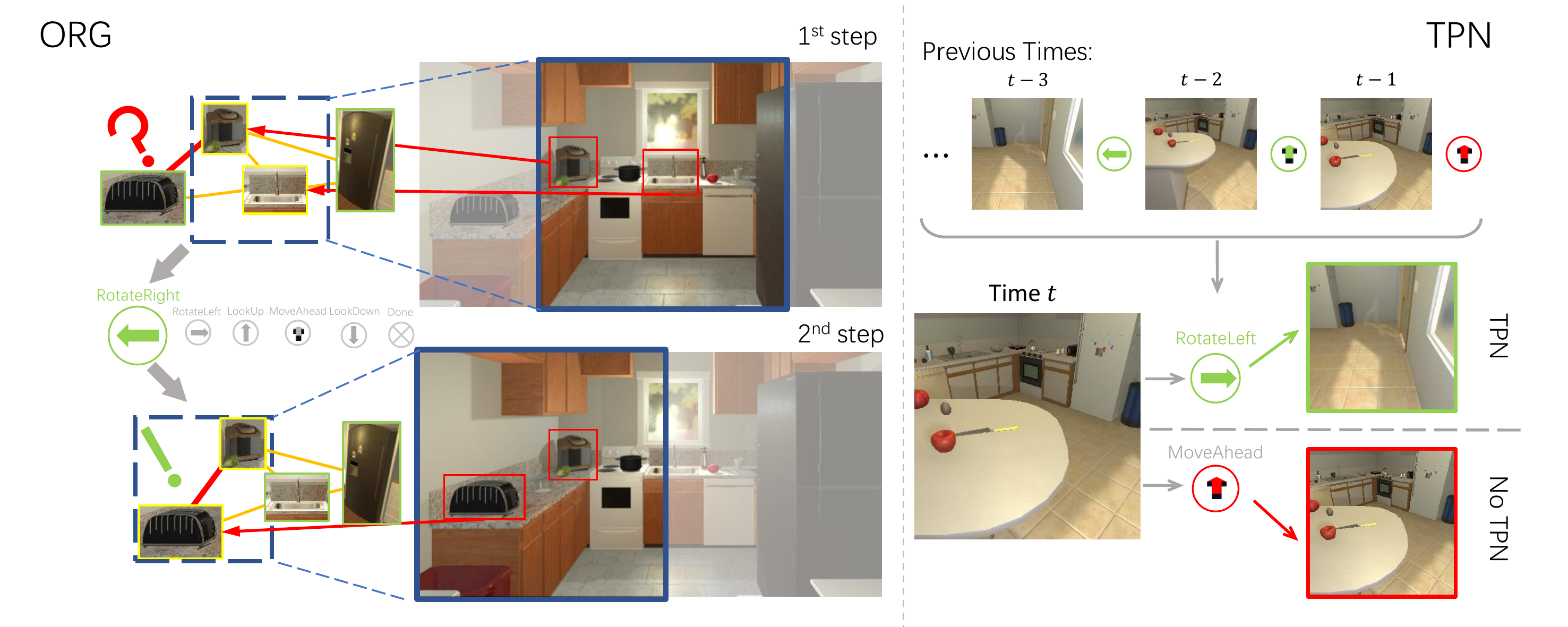}
    \caption{Illustration of our proposed ORG and TPN in unseen testing environments. Right: Looking for a toaster. The agent first sights the coffee machine and our ORG advises that the coffee machine is usually close to the toaster. Given that the agent has not detected the toaster (on the right side of the coffee machine), the agent will turn left to find it. Left: illustration of escaping deadlock by TPN. Based on the current observation, the agent repeats action \texttt{MoveAhead} and falls in deadlock. However, our TPN lets the agent select action \texttt{RotateRight}, thus breaking the deadlock state.}
    \label{fig:introduction}
\end{figure}



Visual navigation aims to steer an agent towards a target object based on its first-view visual observations. To achieve this goal, a mapping from visual observations to agent actions is expected to be established. Thus, representing visual observations and designing navigation policy are the two key components in navigation systems.
For example, to ``grab the TV remote'', an agent needs to know what a remote looks like and then searches it in an environment. 
Due to the small size of the target object, an agent might fail to find it within allowed search steps.
Furthermore, an agent may also fail to move towards the target because of the complexity of the environments. Therefore, learning informative visual representation and failure-aware navigation policy is highly desirable.   

We observe that common objects often exhibit very high concurrence relationships. For instance, cushions often lie on a sofa, or a mouse is next to a laptop. 
The concurrence not only indicates the closeness of object categories but also provides important spatial clues for agents to approach to target objects effectively and efficiently, especially when targets are too small or invisible in the current view, as illustrated in Fig.~\ref{fig:introduction}. Leveraging the concurrence relationships, an agent can narrow down the search area and then find small targets, thus increasing the effectiveness and efficiency of navigation. 

Regarding there are various object categories and different environments, it is difficult to manually design concurrence relationships covering different situations comprehensively~\cite{velivckovic2017graph}. 
Instead, in this paper, we propose an object relation graph (ORG) to learn concurrence relationships among object classes from different environments.
Different from traditional graph convolutional networks (GCN)~\cite{velivckovic2017graph} in which a category adjacent matrix is pre-defined or learned from an external knowledge database, our ORG does not need to resort to external knowledge but learns the category closeness and spatial correlations simultaneously from the object detection information from the training dataset. The object detection also provides stronger association between object concepts and their appearances in comparison to previous works~\cite{wortsman2019learning,yang2018visual} that only employ word embedding to establish the association.
To let an agent focus on moving towards targets without being distracted, we develop a graph attention layer. 
Our graph attention layer emphasizes target related object features while suppressing irrelevant ones via our ORG. In this manner, our extracted local features are more discriminative, thus facilitating object localization.



Due to the complexity of an environment, an agent might fail to reach the target and is stuck in a deadlock state, \eg, repeating the same actions. Only using reinforcement learning in training cannot solve this problem since the reward does not provide explicit guidance of leaving deadlock states to an agent. Thus, an explicit instruction is required to provide when an agent is trapped in the deadlock. Inspired by human trial-and-practice behaviors, we propose a trial-driven imitation learning (IL) supervision to guide the agent with the expert experience to avoid deadlock. 
In this manner, we can continue training our policy network, improving the effectiveness of our navigation policy network.
However, if we clone the expert experience at every step, the policy network will overfit to the seen training environment.

In unseen testing environments, the IL supervision is not available to an agent and it may fall in deadlock in testing.
In order to enable an agent to avoid deadlock states in testing, we develop a memory-augmented tentative policy network (TPN).
Our TPN firstly employs an external memory to record visual representations for detecting deadlock states. When the visual representations are repeated, it implies that an agent may fall in deadlock states. Then, TPN utilizes an internal memory that stores the past state and action pairs to generate explicit instructions for the agent, allowing it to leave deadlock in testing, as visible in Fig.~\ref{fig:introduction}. 
Unlike the work~\cite{wortsman2019learning} that provides a scalar reward at every step in testing, our TPN provides explicit action instructions at failure steps to update our navigation network. Therefore, our method obtains a failure-aware navigation policy.

We adopt the standard A3C architecture~\cite{mnih2016asynchronous} to learn our navigation policy in the artificial environment AI2-Thor \cite{ai2thor}.
Experiments in \emph{unseen} scenes demonstrate that our method achieves superior navigation performance to the baseline methods.
Remarkably, we improve the success rate from 56.4\% to 69.3\% and Success weighted by Path Length (SPL) from 0.319 to 0.394.

Overall, our major contributions are summarized as follows: 
\begin{itemize}
    \item We propose a novel object representation graph (ORG) to learn a category concurrence graph including category closeness and spatial correlations among different classes. Benefiting from our learned ORG, navigation agents are able to find targets more effectively and efficiently. 
    \item We introduce trial-driven imitation learning to provide expert experience to an agent in training, thus preventing the navigation network from being trapped in deadlock and improving its training effectiveness.
    \item To the best of our knowledge, we are the first to propose a memory-augmented tentative policy network (TPN) to provide deadlock breaking policy in the \emph{testing} phase. By exploiting our TPN, an agent is able to notice deadlock states and obtains an escape policy in unseen testing environment.
    \item Experimental results demonstrate that our method significantly improves the baseline visual navigation systems in unseen environments by a large margin of 22.8\% in terms of the success rate.
\end{itemize}

\section{Related Work}

Visual navigation, as a fundamental task in robotic and artificial intelligence, has attracted great attention in the past decades. 
Prior works often require an entire map of an environment before navigation and have been divided into three parts: mapping, localization and path planning.  
The works employ a given map to obviate obstruction~\cite{borenstein1989real,borenstein1991vector} while others use a map for navigation \cite{oriolo1995line}. 
Dissanayke~\etal~\cite{dissanayake2001solution} infer positions from the techniques of simultaneous localization and mapping \cite{dissanayake2001solution} (SLAM). However, a map of an environment is not always available and those methods are not applicable in unseen environments.

Benefitting from the significant progress of the Deep neural networks (DNN), Gupta~\etal~\cite{gupta2017cognitive} introduce cognitive mapping and planning (CMP) to build a map and then plan a route through deep neural network.
Recently, reinforcement learning (RL) based visual navigation approaches aim at taking the current visual observation as input and predicting an action for the next step without intermediate steps, \ie, mapping and planning. 

Mirowski~\etal \cite{mirowski2016learning} adapt two auxiliary tasks, namely predict depth and loop closure classification, to improve navigation performance in complex 3D mazes environment.
The methods~\cite{chen2019learning,fang2019scene} adopt a collision reward and collision detector to avoid collisions.
Several works exploit more information from environments to improve navigation performance. 
Natural-language instruction are available in~\cite{anderson2018vision,wang2019reinforced} to guide the agent actions.
The methods~\cite{sepulveda2018deep,chen2019behavioral,savinov2018semi} propose to use both visual observation features and the topological guidance of scenes.
Furthermore, Kahn~\etal~\cite{kahn2018self} purpose a self-supervised approach to build a model of an environment through reinforcement learning. 
Wu~\etal~\cite{wu2019bayesian} propose a Bayesian relational memory to build room correlations. 
Meanwhile, Shen~\etal \cite{shen2019situational} produce a robust action based on multiple actions from different visual representations.

Recently, target-oriented visual navigation methods have been proposed to search different kinds of object in an environment. 
Zhu~\etal \cite{zhu2017target} employ reinforcement learning to generate an action for the next step based on the current visual observation and a given destination image instead of a specific target class.
Mousavian~\etal~\cite{mousavian2019visual} fuse semantic segmentation and detection masks and then feed the fused features into their policy network for navigation. 
Furthermore, Wortsman \etal \cite{wortsman2019learning} adopt Glove embedding to represent target objects and 
a network to simulate the reward function in reinforcement learning for navigation in unseen environments.
Similar to the works~\cite{fang2019scene,battaglia2016interaction}, Yang~\etal~\cite{yang2018visual} propose a graph convolutional network \cite{velivckovic2017graph} to exploit relationships among object categories, but they need to resort to an external knowledge database and do not explore the category spatial correlations. 
However, those works may suffer semantic ambiguity or non-discriminative representations of the visual information, and thus navigate an agent to contextually similar objects instead of targets or fail to recognize targets.
In contrast, our method exploits the detection results and thus significantly alleviates the semantic ambiguity. Moreover, our memory augmented TPN is the first attempt to enable an agent to escape from deadlock states in the testing phase among reinforcement learning based navigation systems. 

\begin{figure}[t]
    \centering
    \includegraphics[width=0.85\linewidth]{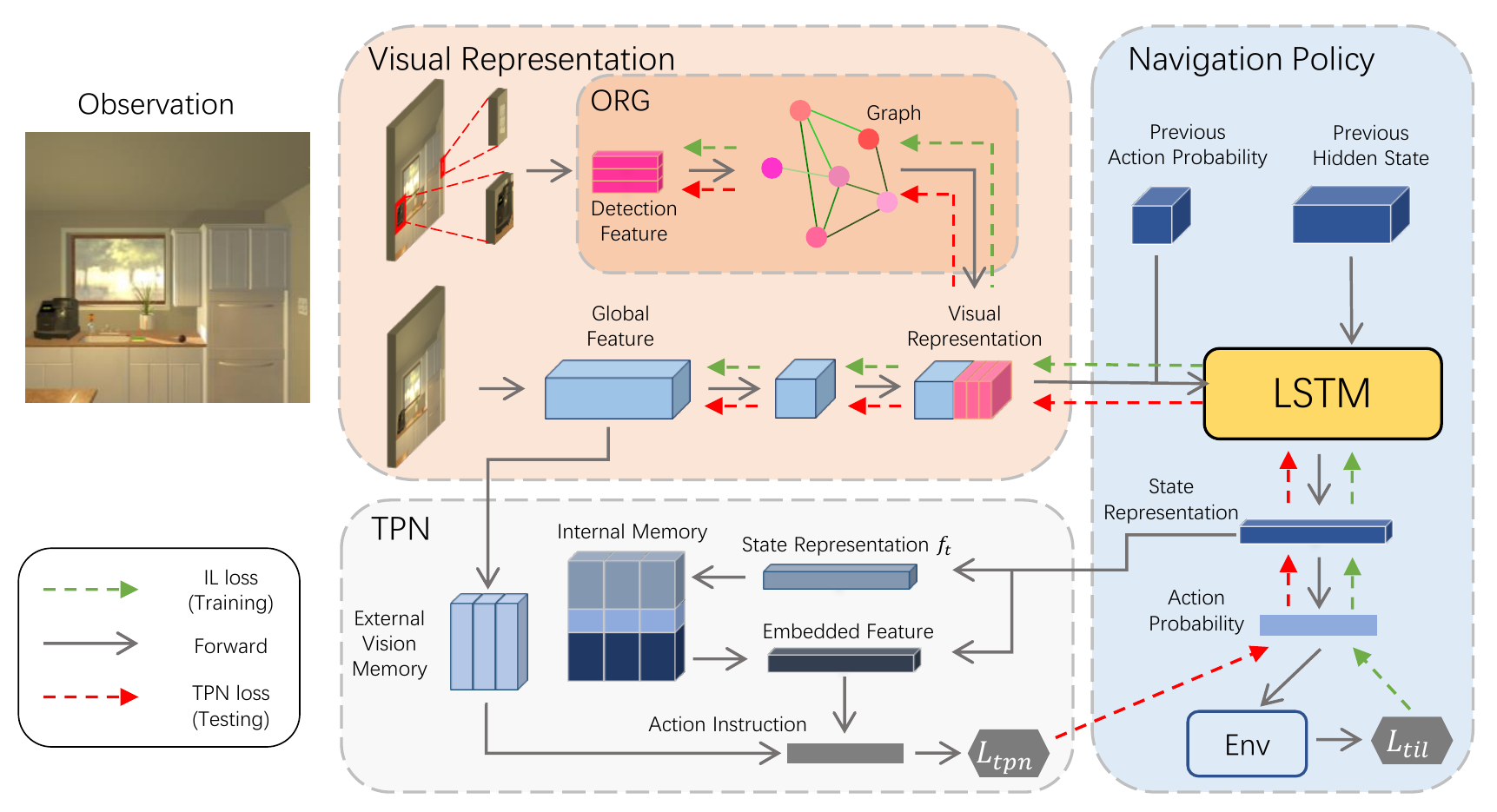}
    \caption{\textbf{Overview of our proposed framework.} The visual representation is combined by the global feature and local feature encoded by our ORG. The navigation network adopts A3C model but trained with both the reinforcement learning reward and our trial-driven IL supervision. TPN is trained in the deadlock states. In training TPN, our navigation network is fixed. In testing, TPN updates our navigation network.
    }
    \label{fig:pipeline}
\end{figure}

\section{Proposed Method}
Our goal is to introduce an informative visual representation and a failure-aware navigation policy for a target-driven visual navigation system. 
To achieve this goal, our navigation system contains three major components, as illustrated in Fig.~\ref{fig:pipeline}:
(i) learning visual representation from RGB observations; In this component, we introduce an object representation graph (ORG) to extract informative visual representation for objects of interest.
(ii) learning navigation policy based on our visual representation; To prevent an agent from being trapped in local minima, such as deadlock states, in training, we propose trial-driven imitation learning.
(iii) learning a tentative policy network; This allows an agent to receive policy instruction in order to break deadlock during testing.



\subsection{Task Definition}
Given a target object category, \eg, remote control, our task is to navigate an agent to an instance of this class using visual information.
During navigation, RGB images in an egocentric view are the only available source for an agent and the agent predicts its actions based on the current view. 
Information about the entire environment, \ie topological map and 3D meshes, is not available to the agent. 
Meanwhile, an environment is divided into grids and each grid node represents one unique state in the environment.
In all the environments, an agent is able to move between nodes with 6 different actions, including \texttt{MoveAhead, RotateLeft, RotateRight, LookUp, LookDown, Done}. 

One successful episode is defined as: an agent selects the termination action~\texttt{Done} when the distance between the agent and the target is less than a threshold (\ie, 1.5 meters) and the target is in its field of view.
If a termination action is executed at any other time, the agent fails and the episode ends.
 
At the beginning of each episode, an agent is given a random target class word $T\in\{ \texttt{Sink}, \dots , \texttt{Microwave} \}$ and starts from a random state $s=\{x, y, \theta_r, \theta_h \}$ in a random room to maintain the uniqueness of each episode, where $x$ and $y$ represent the position of the agent, $\theta_r$ and $\theta_h$ indicate the point of view of the agent. 
At each timestamp $t$, an agent captures the current observation $O_t$ in the first-person perspective. Based on the visual representation extracted from $O_t$ and the $T$, the agent generates a policy $\pi(a_t | O_t, T)$, where $a_t$ represents the distribution of actions, and the action with the highest probability is selected for the next step. The agent will continue moving until the action~\texttt{Done} is issued. 

\subsection{Object Representation Graph}
\begin{figure}[t]
    \centering
    \includegraphics[width=0.9\textwidth]{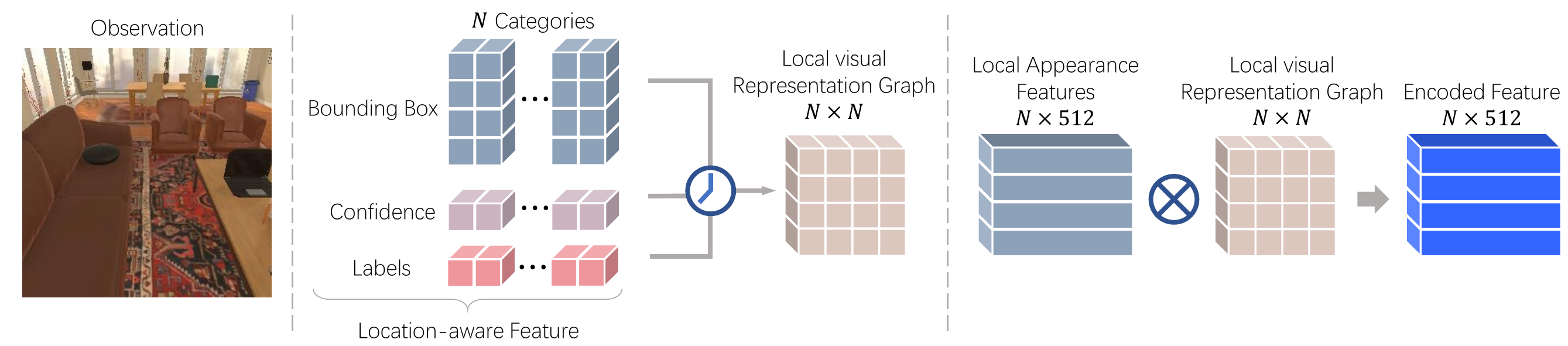}
    \caption{\textbf{Illustration of object representation graph.} The agent extracts the LAF, including bounding boxes, confidences and the target label, from the current observation. Then the agent adopts the LAF to generate the ORG. To emphasize the region of interest, we employ ORG as an attention map to encode the local appearance features.}
    \label{fig:graph}
\end{figure}

Regarding the agent observes an environment in an egocentric view instead of a bird's-eye view, how to design an effective exploration method plays a critical role in visual navigation.
Inspired by the human searching behaviors, we aim to fully explore the relationship among categories as well as their spatial correlations for navigation. 
Therefore, we introduce an object representation graph network to explore such concurrence information.

\noindent{\bf{Detection and location-aware feature.}}
In order to learn the relationship among classes and their spatial correlations, we need to find all the objects in an image first. 
Thus, we train an object detector, \ie, Faster RCNN \cite{ren2015faster}, to achieve this goal. 
Given an input image, we first perform object detection to localize all the objects of interest. 
If there are multiple instances of an object class, we only choose the one with the highest confidence score. 
We record the bounding box positions and detection confidence for each category and then concatenate them as our local detection feature. 
It is likely that some category objects do not appear in the current view. 
Therefore, we record the bounding box positions and confidence of those categories as 0.
In order to provide the target information during navigation, we concatenate a one-hot encoded target vector with our local detection feature as our location-aware feature (LAF), as seen in Fig.~\ref{fig:graph}.
Moreover, we extract not only the location feature but also appearance feature for the object.
We project the bounding boxes to the same layer in the backbone network of the detector and then extract our location-aware appearance features, as seen in Fig.~\ref{fig:graph}. Due to the small resolution of input images~\cite{ai2thor}, we extract appearance features from the second ResBlock layer in the backbone network to preserve spatial details of local regions. 

\noindent{\bf{Learning object representation graph.}}
After obtaining our extracted LAF, we introduce our graph convolutional network to learn our object representation graph (ORG). 
We first define a graph by $G=(N,A)$, where $N$ and $A$ denote the nodes and the edges between nodes respectively. 
To be specific, each node $n\in N$ denotes the concatenated vector of the bounding box position, confidence and label (see Fig.~\ref{fig:graph}), and each edge $a\in A$ denotes the relationships among different classes.
Our graph convolutional network (GCN) takes all the nodes as inputs $X \in \mathbb{R}^{|N|\times D}$ and then embeds each input node by a matrix $W \in \mathbb{R}^{D\times N}$, where $D$ indicates the dimension of our LAF. 
After encoding each node, our GCN embeds, regarded as convolution, all the nodes according to the adjacent relationship $A \in \mathbb{R}^{|N|\times N}$ and outputs a new encoding $Z \in \mathbb{R}^{|N|\times N}$. 
Our graph convolutional layer is expressed as:
\begin{equation}
    Z = f(A\cdot X\cdot W),
\end{equation}
where $f(\cdot)$ denotes the ReLU activation function. Different from traditional GCNs in which an adjacent matrix $A$ is often pre-defined, our ORG network learns the node embedding $W$ as well as the adjacent matrix $A$. The process of learning $A$ actually can be regarded as encoding the spatial correlations among categories as well as their relationships since $A$ encodes the embedded LAF across different categories.
The output $Z$ (\ie, ORG) encodes the location information among objects and their closeness. 
Moreover, since our object representation graph is learned in accordance with environments rather than a graph learned from external databases, such as FastText \cite{joulin2016bag}, our ORG is able to adapt to different environments. 

\noindent{\bf{Graph attention layer.}}
To let the agent focus on moving towards the target or the areas where the target is likely placed, we adopt an attention mechanism in our network.
Specifically, we employ our $Z$ as our attention map to the location-aware appearance feature. 
Denote our location-aware appearance feature as $F\in\mathbb{R}^{|N|\times d}$, where $d$ represents the dimension of our location-aware appearance feature. Our graph attention layer is expressed as:
\begin{equation}
    \hat{F} = f(Z\cdot F),  
\end{equation}
where $\hat{F}\in\mathbb{R}^{|N|\times d}$ is our attended location-aware appearance feature. Note that, there is no learnable parameters in our graph attention layer. 
Then we concatenate our attentive location-aware appearance feature with LAF for explicit target location information.

Another advantage of our graph attention module is that our concatenated location-aware appearance feature is more robust than $X$. 
For instance, when our detector fails to localize targets or produces false positives, our model is still able to exploit target related information for navigation. 
In contrast, $X$ does not provide such concurrence relationships among objects and an agent needs more steps to re-discover target objects. 
This also implies that using concatenated location-aware appearance feature we can achieve a more efficient navigation system. When the category relationships may not follow our learned ORG, our LAF (from our detector) is still valid for navigation and ensures the effectiveness of our navigation system in those cases.

\subsection{Navigation Driven by Visual Features}

Besides the task-specific visual representations, such as our concatenated location-aware appearance feature, an agent requires a global feature to describe the surroundings of an environment.
Similar to~\cite{wortsman2019learning}, we employ ResNet18~\cite{he2016deep} pretrained on ImageNet~\cite{deng2009imagenet} to extract the global feature of the current view. 
We then fuse the global visual feature as well as our concatenated location-aware appearance feature as our final visual representation. 

We adopt the standard Asynchronous Advantage Actor-Critic (A3C) architecture~\cite{mnih2016asynchronous} to predict policy at each step. 
The input of our A3C model is the concatenation of our visual representation, the previous action and state embedding. 
Recall that the representation of previous actions and state embedding are feature vectors while our visual representation is a feature volume. 
Thus, we repeat them along the spatial dimensions so as to fit the size of our visual representation, and then feed the concatenated features to our A3C model. Our A3C model produces two outputs, \ie, policy and value. 
We sample the action from the predicted policy with the highest probability and the output value is used to train our policy network.

\subsection{Trial-driven Imitation Learning}
\label{sec:imitation_learning}
Concerning the complexity of the simulation environment, an agent may be trapped into deadlock states. 
Since the reinforcement reward cannot provide detailed action instruction for deadlock breaking, agents are difficult to learn escape policy without explicit supervision. 
Therefore, we propose trial-driven imitation learning (IL) to advise agents through explicit action instructions.
To learn optimal action instructions, we employ expert experience acting as the policy guidance for an agent.
We use Dijkstra's Shortest Path First algorithm to generate the expert experience.
Under the supervision of policy guidance, an agent is able to imitate the optimal deadlock breaking solution. 
The IL loss $L_{il}$ is given by the cross-entropy $L_{il} = CE(a_t, \hat{a})$, where $a_t$ is the action predicted by our navigation policy, $\hat{a}$ represents the action instruction and $CE$ indicates the cross-entropy loss. The total training loss $L$ for training our navigation policy network is formulated as:
\begin{equation}
    L = L_{nav} + L_{il},
\end{equation}
where $L_{nav}$ represents our navigation loss from reinforcement learning.

Due to the limited training data, imitation learning may lead navigation policy to overfitting seen environments after millions of episode training.
In order to maintain the generalization ability of agents to unseen environments, we need to balance imitation learning and reinforcement learning.
Inspired by human trial-and-practice behaviors, we utilize the policy guidance in deadlock states instead of every state. 
In doing so, we can continue the episode instead of staying in deadlock states till termination, thus improving our training effectiveness.
For instance, when the target object is in the corner of the room, using our imitate learning supervision, our agent is able to escape from deadlock and reach the target after a few turns. In contrast, without IL supervision, the agent traps in a position far away from the target till the episode ends.

\begin{figure}[t]
    \centering
    \includegraphics[width=0.9\textwidth]{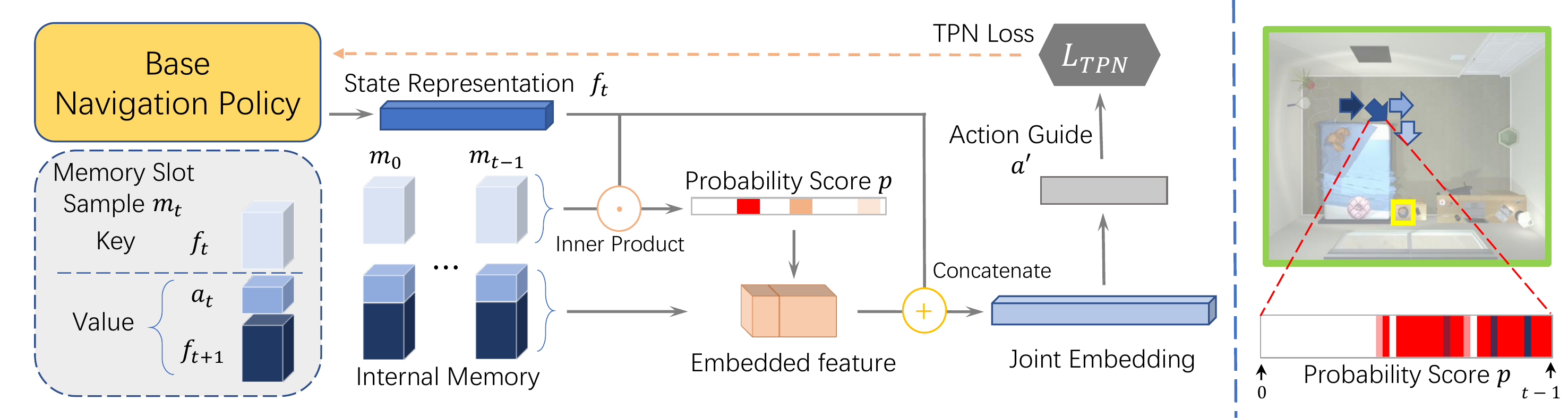}
    \caption{\textbf{Illustration of memory-argumented tentative policy network.} Right: Agent compares the state representation with the key of the internal memory and then generates a weight to encode the past state and action pairs. The embedded feature and the state are concatenated to learning an action for breaking deadlock. The supervision of TPN comes from expert experience of IL. Left: visualization of the probability score $p$ when TPN guides the agent to escape from the deadlock states. The darker color indicates which previous state will be more likely used for learning policy.}
    \label{fig:tpn}
\end{figure}

\subsection{Memory-Augmented Tentative Policy Network}
{\bf{External Memory.}} Unlike in the training stage, instructions from an environment, such as expert experience and validation information of actions, are not available to agents in testing.
Therefore, we propose a memory-augmented tentative policy network to assist an agent to break deadlock.
In order to detect the deadlock states, we employ an external vision memory.
Our external vision memory is designed to record visual features from an off-the-shelf feature extractor.
Once there is at least one visual feature as the same as those recorded in memory, we assume an agent is stuck in deadlock states.
Denoted an episode by $\{s_0, s_1, \dots, s_t\}$, where $s_t$ represents a states of an agent at time $t$.
We define $s_t$ as a deadlock if the visual features extracted from $s_t$ and another previous state $s_t'$ are similar. 

{\bf{Internal Memory.}} In order to capture long-term dependencies and generate instructions based on past states, we present an internal state memory.
Different from the external vision memory, our internal state memory is designed to store state and action pairs.
Each memory slot $m_t$ at time $t$ includes two components: 
(i) the state representation $f_{t}$ at time $t$ serving as a key of $m_t$;
(ii) both the action distribution $a_t$ at time $t$ and the transformed state representation $f_{t+1}$ at time $t+1$ serving as value of $m_t$.
In each step, a newly-generated memory slot will be inserted at the end of the internal state memory.

{\bf{Tentative Policy Network (TPN).}} To fully utilize the previous adventures, our TPN first employs a soft attention mechanism to generate pseudo expert experience from our internal memory.
Given the preceding actions and state transformation, TPN computes the probability score $p$ between keys $k$ of each memory slot and current state representation $f_t$ by taking the inner product followed by a softmax,
\begin{equation}
    p = \sigma(f_t^{T} \cdot k),
\end{equation}
where $\sigma(x_i) = \frac{e^{x_i}}{\sum_{j}{e^{x_j}}}$.
Then, the embedded memory feature is the weighted sum over the value of memory slots by the probability score, as illustrated in Fig.~\ref{fig:tpn}.

To obtain informative representation, we concatenate the embedded memory feature with the current state and then encode them as a joint feature embedding.
After that, TPN exploits the joint feature embedding to generate the action guidance for our base navigation policy network.
In this manner, TPN will provide deadlock breaking policy based on previous action and state pairs. 

In order to train TPN, we use our trained base navigation network to navigate in an environment. When the agent falls into deadlock, we use our imitation learning supervision to train our TPN. 
In this fashion, our TPN learns how to provide deadlock breaking actions in the deadlock situation.
In testing, our TPN is fixed and an agent will update its base navigation policy by the cross-entropy $L_{tpn} = CE(a_t, a')$, where $a_t$ is the action predicted by the base navigation policy and $a'$ indicates the action from the expert experience.
Overall, our trial-driven imitation learning supervision and TPN facilitate an agent to establish a failure-aware navigation policy in both training and testing. 

\subsection{Training Details}
We train our model in two stages:
(i) training navigation policy for 6M episodes in total with 12 asynchronous agents; In this stage, we use trial-driven imitation learning and reinforcement learning rewards as our objective.
(ii) training our TPN for 2M episodes in total with 12 asynchronous agents; We select the navigation model performing the best on the \textit{validation set} in terms of success rate as the fixed backbone to train our TPN.
Both training stages are performed on the training set.
Similar to \cite{wortsman2019learning}, in learning navigation policy we penalize each action step with $-0.001$. When an agent reaches a target and sends the termination signal \texttt{Done}, we will reward the agent with a large value 5. 
In our experiments, we employ Adam optimizer~\cite{kingma2014adam} to update the parameters of our networks with a learning rate $10^{-4}$. 

We employ Faster RCNN as our detector and re-train it on the training dataset, (\ie, AI2-Thor environment \cite{ai2thor}). We employ half of the training dataset and data augmentation to train our detector to avoid overfitting. We will release our training protocols and codes for reproducibility.  

\section{Experiments}
\subsection{Dataset and Evaluation}
\textbf{Dataset.}  
We choose AI2-Thor \cite{ai2thor} environment to evaluate our method. 
AI2thor dataset contains four types of scenes, including kitchen, living room, bedroom and bathroom, and each scene includes 30 rooms, where each room is unique in terms of furniture placement and item types. 
Following ~\cite{wortsman2019learning}, we select 22 categories from those four types of scenes. In each scene, there are more than four target classes, and an agent randomly starts navigation at over 2000 states. 

\textbf{Evaluation.}  
We use the success rate and Success Weighted by Path Length (SPL) for performance evaluation. 
The success rate measures the effectiveness of trajectories and is formulated as $\frac{1}{N} \sum_{n=0}^{N} S_{n}$, where $N$ stands for the total number of episodes, and $S_{n}$ is the binary indicator of $n$-th episode. 
SPL evaluates the efficiency of the model through $\frac{1}{N} \sum_{n=0}^{N} \frac{Len_{n}}{max(Len_{n}, Len_{opt})}$, where $Len_{n}$ and $Len_{opt}$ represent the length of the $n$-th episode and its optimal path, respectively.

\subsection{Task Setup and Comparison Methods}
We use the evaluation protocol in \cite{wortsman2019learning}. 
To ensure the generalization of our method, there is no overlap between our training rooms and testing ones.
We select 25 out of 30 rooms per scene as the training and validation set. 
We test our method only in the remaining 20 {\bf unseen} rooms.
During the evaluation, each model performs 250 episodes per scene from the validation set. The model with the highest success rate will be performed on the test set as the reported results. 


\textbf{Baseline.} We feed the detection results to A3C for navigation as our baseline, on top of which we build our model.

\textbf{Random policy.} An agent navigates based on a uniform action probability. Thus, the agent will randomly walk in the scene or randomly stop.

\textbf{Scene Priors (SP).} \cite{yang2018visual} exploits a category relation graph learned from an external database, FastText~\cite{joulin2016bag}. We replace its original WE with detection results for fair comparison, dubbed {\bf D-SP}.

\textbf{Word Embedding (WE).} An agent uses Glove embedding to associate target concepts and appearances. 

\textbf{Self-adaptive Visual Navigation method (SAVN).} \cite{wortsman2019learning} introduces a meta reinforcement learning method in unseen environments. Furthermore, SAVN employs WE to associate target concepts and appearances. We replace its original WE with detection results to achieve a stronger baseline, dubbed {\bf D-SAVN}.

\subsection{Results}
{\bf Quantitative Results.}
We demonstrate the results of four comparison methods and our baseline model in Table~\ref{tab:results_baselines}. 
For fair comparisons, we also follow the setup and protocols in \cite{wortsman2019learning} when measuring the performance of our method.

\begin{table}[t]\renewcommand{\arraystretch}{0.8}
    \centering
    \caption{Comparisons of navigation results. We report the success rate (\%), denoted by Success, and SPL. $L>5$ indicates the optimal path is larger than 5 steps}
    \setlength{\tabcolsep}{1.5mm}{
        \begin{tabular}{l|cc|cc}
            \toprule
            \multirow{2}{*}{{\scriptsize Method}}                   & \multicolumn{2}{c|}{{\scriptsize ALL}}                            & \multicolumn{2}{c}{{\scriptsize$L \ge 5$}}                            \\ \cline{2-5}
                                                                    & {\scriptsize Success}           & {\scriptsize SPL}               & {\scriptsize Success}             & {\scriptsize SPL}                 \\
            \hline \hline
            {\scriptsize Random}                                    & {\scriptsize 8.0}               & {\scriptsize 0.036}             & {\scriptsize 0.3}                 & {\scriptsize 0.001}               \\
            {\scriptsize WE}                                        & {\scriptsize 33.0}              & {\scriptsize 0.147}             & {\scriptsize 21.4}                & {\scriptsize 0.117}               \\
            {\scriptsize SP~\cite{yang2018visual}}                  & {\scriptsize 35.1}              & {\scriptsize 0.155}             & {\scriptsize 22.2}                & {\scriptsize 0.114}               \\
            {\scriptsize D-SP~\cite{yang2018visual}}                & {\scriptsize 59.6}              & {\scriptsize 0.303}             & {\scriptsize 47.9}                & {\scriptsize 0.273}               \\
            {\scriptsize SAVN~\cite{wortsman2019learning}}          & {\scriptsize 40.8}              & {\scriptsize 0.161}             & {\scriptsize 28.7}                & {\scriptsize 0.139}               \\
            {\scriptsize D-SAVN~\cite{wortsman2019learning}}        & {\scriptsize 62.3}              & {\scriptsize 0.264}             & {\scriptsize 53.3}                & {\scriptsize 0.254}               \\ \hline
            {\scriptsize Baseline}                                  & {\scriptsize 56.4}              & {\scriptsize 0.319}             & {\scriptsize 42.5}                & {\scriptsize 0.270}               \\
            {\scriptsize Baseline + TPN}                            & {\scriptsize 58.7}              & {\scriptsize 0.316}             & {\scriptsize 45.8}                & {\scriptsize 0.274}               \\ 
            {\scriptsize Baseline + IL}                             & {\scriptsize 63.6}              & {\scriptsize 0.354}             & {\scriptsize 52.8}                & {\scriptsize 0.326}               \\
            {\scriptsize Baseline + ORG}                            & {\scriptsize 65.3}              & {\scriptsize 0.375}             & {\scriptsize 54.8}                & {\scriptsize 0.361}               \\ \hline
            {\scriptsize \textbf{Ours (TPN+ORG+IL)}}                & {\scriptsize \textbf{69.3}}     & {\scriptsize \textbf{0.394}}    & {\scriptsize \textbf{60.7}}       & {\scriptsize \textbf{0.386}}      \\ \bottomrule
        \end{tabular}
    }
    \label{tab:results_baselines}
\end{table}

\begin{figure}[t]
    \centering
    \includegraphics[width=0.7\linewidth]{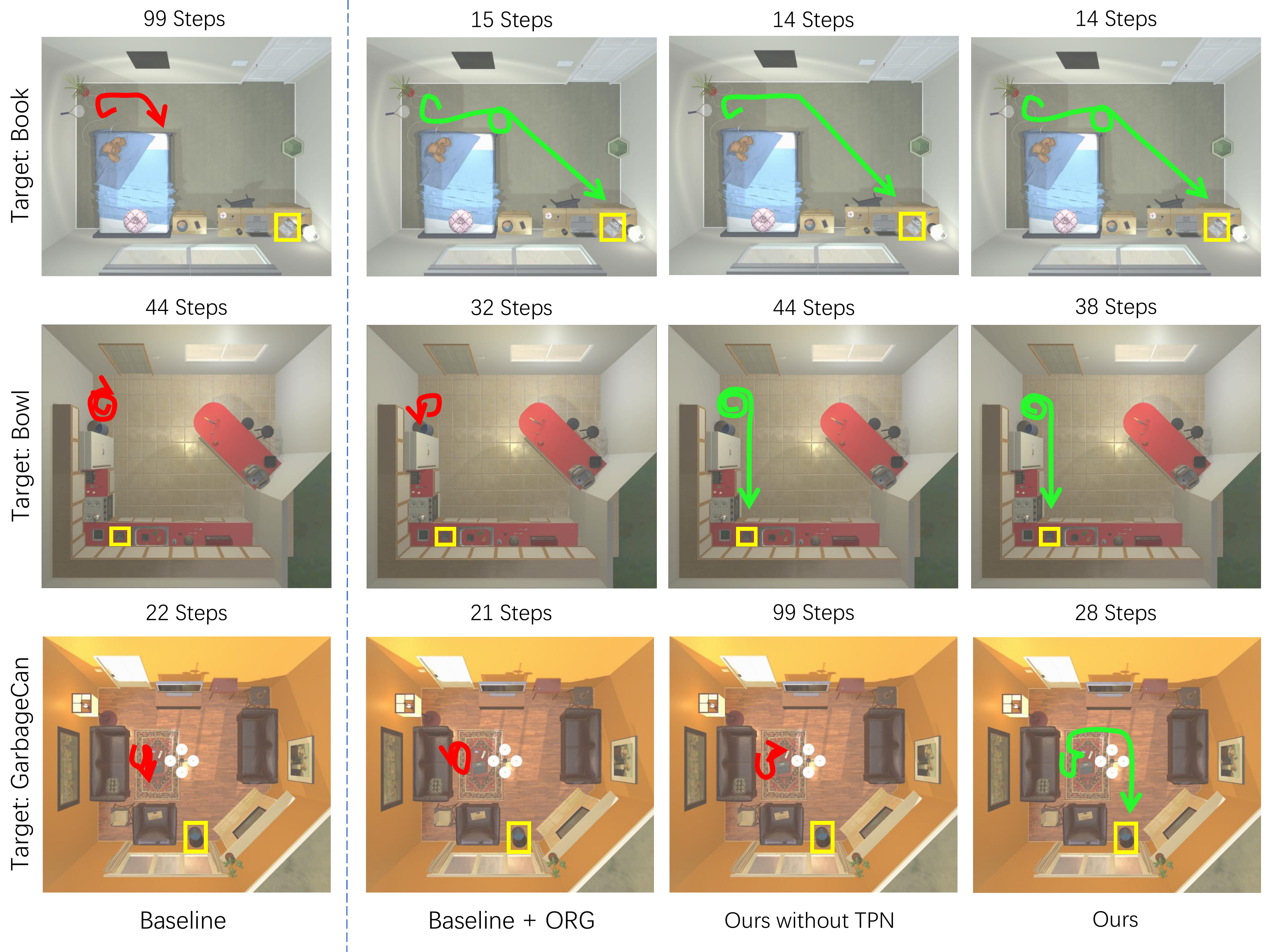}
    \caption{\textbf{Visual results of different models in testing environments.} We compare our proposed model with our proposed model without TPN and our baseline with/without ORG.
    The target objects are highlighted by the yellow bounding boxes. Green and red lines represent success cases and failure cases, respectively.
    First row: the target is \textit{book}. Second row: the target is \textit{bowl}. Third row: the target is \textit{Garbage Can}.
    }
    \label{fig:case_study}
\end{figure}

As indicated in Table~\ref{tab:results_baselines}, our method outperforms our baseline significantly in terms of the success rate and SPL. 
Meanwhile, each module of our method is able to improve navigation performance.
Since the baseline does not exploit category concurrence relation, it needs to search the target object only based on detection. This experiment indicates that our ORG encodes informative visual representation for agents, thus significantly increasing efficiency and effectiveness of navigation.
Furthermore, both our trial-driven imitation learning and TPN are able to predict advisable instructions to guide agent escape from local minima, and thus those two models achieve better performance than our baseline. Note that, our baseline does not have any mechanism to avoid deadlock states.

SP~\cite{yang2018visual} also aims at utilizing category relationships and leverages external knowledge to encode the category relationship.
However, SP also employs WE to associate object appearances and concepts, while our ORG encodes object locations and appearances directly from our detector. 
Therefore, our method achieves superior performance to SP on both the success rate and SPL. 
Unlike D-SP that concatenates a graph representation with detection features from different modalities, our model fuses detection results via a learned graph and thus achieves better performance.

Although state-of-the-art model SAVN employs meta reinforcement learning to improve navigation performance, SAVN uses word embedding~\cite{pennington2014glove} to represent targets, thus suffering the ambiguity when objects often appear together, such as a TV set and a remote.
We replace the word embedding with our detection module, named D-SAVN.
The experiment indicates that the detection information significantly improves the performance of SAVN. 
Compared to D-SAVN that improves navigation effectiveness by simulating a reward in testing, our model explicitly provides instructions to escape from deadlock states and thus achieves better performance on both metrics, as indicated in Table~\ref{tab:results_baselines}.

{\bf{Case Study.}}
Fig.~\ref{fig:case_study} illustrates trajectories of three navigation tasks proceeded by four models, \ie, the baseline, the baseline with ORG, our model without TPN and our full model, in \emph{unseen} testing environments.

In the first case, the baseline fails to find the target and is stuck in the environment, since it reaches the maximum step limit \ie, 99 steps. 
On the contrary, the baseline with ORG finds the target object successfully. This implies that using our ORG, we can improve the navigation effectiveness.  Moreover, the navigation system with ORG only uses 15 steps to localize the object. This indicates our ORG improves the navigation efficiency. 

In the second case, the baseline and the baseline with ORG repeat the same actions until the agents terminate. 
It can be seen that both the baseline and the baseline with ORG are trapped in deadlock states. In contrast, the navigation system trained with IL supervision overcomes the deadlock and reaches the target. This demonstrates the importance of our trial-driven IL supervision.  
Furthermore, our model escapes the deadlock using the least steps, demonstrating TPN improves the navigation effectiveness in testing. 


In the third case, the environment is more complicated. It can be seen that the baseline with and without ORG both fail to find the target since they are trapped in deadlock states. As seen in the third column, the model trained with IL manages to escape the deadlock, but reaches the maximum step limit and fails for the lack of explicit instruction in testing. Benefiting from TPN, our model leaves the deadlock state and successfully localizes the target. This demonstrates that TPN is very helpful for breaking the deadlock states in testing and improves the navigation effectiveness.

\subsection{Ablation Study}
\begin{table}[t]\renewcommand{\arraystretch}{0.8}
    \centering
    \caption{Impacts of different components on navigation performances}
    \setlength{\tabcolsep}{1mm}{
        \begin{tabular}{cc|ccc|cc|cc|c}
        \toprule
            \multicolumn{2}{c|}{\multirow{2}{*}{{\scriptsize Method}}}              & \multirow{2}{*}{{\scriptsize w/o IL}}         & \multirow{2}{*}{{\scriptsize w/o ORG}}        & \multirow{2}{*}{{\scriptsize w/o TPN}}        & \multicolumn{2}{c|}{{\scriptsize IL}}           & \multicolumn{2}{c|}{{\scriptsize TPN}}        &  \multirow{2}{*}{{\scriptsize Ours}}       \\ \cline{6-9}
            &                                                       &                           &                           &                           & {\scriptsize failed}        & {\scriptsize all}           & {\scriptsize all}       & {\scriptsize random}        &                                        \\ \hline \hline
            \multirow{2}{*}{{\scriptsize ALL}}    & \multicolumn{1}{|c|}{{\scriptsize Success}} & {\scriptsize 66.8\%}                    & {\scriptsize 67.5\%}                    & {\scriptsize 66.6\%}                    & {\scriptsize 63.6\%}        & {\scriptsize 47.7\%}        & {\scriptsize 66.2\%}    & {\scriptsize 62.3\%}        & {\scriptsize \textbf{69.3\%}}             \\
            & \multicolumn{1}{|c|}{{\scriptsize SPL}}                             & {\scriptsize 0.375}                     & {\scriptsize 0.345}                     & {\scriptsize 0.387}                     & {\scriptsize 0.354}         & {\scriptsize 0.284}         & {\scriptsize 0.325}     & {\scriptsize 0.315}         & {\scriptsize \textbf{0.394}}              \\ \hline
            \multirow{2}{*}{$L\ge5$}&\multicolumn{1}{|c|}{{\scriptsize Success}}  & {\scriptsize 57.2\%}                   & {\scriptsize 57.8\%}                   & {\scriptsize 57.4\%}                    & {\scriptsize 52.8\%}        & {\scriptsize 35.3\%}        & {\scriptsize 56.4\%}    & {\scriptsize 49.5\%}        & {\scriptsize \textbf{60.7\%}}             \\
            & \multicolumn{1}{|c|}{{\scriptsize SPL}}                             & {\scriptsize 0.364}                     & {\scriptsize 0.327}                     & {\scriptsize 0.374}                     & {\scriptsize 0.326}         & {\scriptsize 0.218}         & {\scriptsize 0.295}     & {\scriptsize 0.266}         & {\scriptsize \textbf{0.386}}              \\ \bottomrule
        \end{tabular}
    }
    \label{tab:ablation_study}
\end{table}

Our method has three major contributions, \ie ORG, trial-driven IL supervision and TPN. We dissect their impacts as follows.

\textbf{Impact of trial-driven IL.}
As seen in Table~\ref{tab:ablation_study}, our trial-driven IL improves the navigation policy compared to the model without using IL supervision to train our navigation network (w/o IL).
This manifests that involving clear action guidance in deadlock states improves the navigation results compare to navigation rewards.
Furthermore, to study the influence of IL on generalization, we train our baseline model with IL at every step, marked all in IL.
Table~\ref{tab:ablation_study} implies that providing IL supervision at every step will overfit to the training data, thus undermining the generalization of navigation systems to unseen environments.

\textbf{Impact of ORG.}
Our ORG improves the performance of navigation systems compared with the model without ORG (w/o ORG), as indicated in Tab.~\ref{tab:ablation_study}. Note that the significant SPL improvements demonstrate ORG improves the efficiency of navigation systems.

\textbf{Impact of TPN.}
As indicated in Tab.~\ref{tab:ablation_study}, our TPN improves the success rate and SPL for our navigation system compared to the model without TPN (w/o TPN).
It implies that our TPN helps agents to break deadlock in testing.
Since our TPN focuses on learning deadlock avoidance policy, using TPN updates our based navigation network at every state will harm the navigation performance, marked all in TPN. 
As seen in Tab.~\ref{tab:ablation_study}, using random actions to solve deadlock states (random in TPN) suffers performance degradation.
This indicates that our TPN predicts reasonable escape policy rather than random walking. 


\section{Conclusions}
In this paper, we proposed an effective and robust target-driven visual navigation system. 
Benefiting from our proposed object representation graph, our navigation agent can localize targets effectively and efficiently even when targets are invisible in the current view. 
Furthermore, our proposed trial-driven imitation learning empowers our agent to escape from deadlock states in training, while our tentative policy network allows our navigation system to leave deadlock states in unseen testing environments, thus further promoting navigation effectiveness and achieving better navigation performance.
Experiments demonstrate that our method achieves state-of-the-art performance.

\noindent \textbf{Acknowledgements.} Dr. Liang Zheng is the recipient of Australian Research Council Discovery Early Career Award (DE200101283) funded by the Australian Government. This research was also supported by the Australia Research Council Centre of Excellence for Robotics Vision (CE140100016).



\clearpage
%
%
\bibliographystyle{splncs04}
\bibliography{egbib}
\end{document}